\documentclass[conference]{IEEEtran}
\IEEEoverridecommandlockouts
\usepackage{cite}
\usepackage{amsmath,amssymb,amsfonts}
\usepackage{algorithmic}
\usepackage{graphicx}
\usepackage{textcomp}
\usepackage{cuted}

\begin{document}
\title{Sequence to sequence deep learning models for solar irradiation forecasting
}

\author{\IEEEauthorblockN{Bhaskar Pratim Mukhoty, Vikas Maurya, Sandeep Kumar Shukla}
	\IEEEauthorblockA{\textit{Department of CSE} \\
		\textit{Indian Institute of Technology Kanpur}\\
		Kanpur, India \\
		\{bhaskarm, vikasmr, sandeeps\}@cse.iitk.ac.in}
}

\maketitle

\begin{abstract}
The energy output a photo voltaic(PV) panel is a function of solar irradiation and weather parameters like temperature and wind speed etc. A general measure for solar irradiation called Global Horizontal Irradiance (GHI), customarily reported in Watt/meter$^2$, is a generic indicator for this intermittent energy resource. An accurate prediction of GHI is necessary for reliable grid integration of the renewable as well as for power market trading. While some machine learning techniques are well introduced along with the traditional time-series forecasting techniques, deep-learning techniques remains less explored for the task at hand. In this paper we give deep learning models suitable for sequence to sequence prediction of GHI. The deep learning models are reported for short-term forecasting $\{1-24\}$ hour along with the state-of-the art techniques like Gradient Boosted Regression Trees(GBRT) and Feed Forward Neural Networks(FFNN).

We have checked that spatio-temporal features like wind direction, wind speed and GHI of neighboring location improves the prediction accuracy of the deep learning models significantly. Among the various sequence-to-sequence encoder-decoder models LSTM performed superior, handling short-comings of the state-of-the-art techniques.
\end{abstract}


\section{Introduction}
\label{sec:introduction}
Renewable energy is gradually becoming the future of power resource driven by reasons like climate change, energy independence and security. In the year 2017, the global power share of renewables were $24.3\%$ after experiencing a growth of $17\%$ which remained approximately same over the last decade.\cite{BP}

Among the renewables solar irradiation is a globally popular source of energy, possibly due to it's very low carbon foot-print, abundant availability and rapid advancement of photo-voltaic (PV) technology. \cite{kabir2018solar,Report_2011}. Several fronts of research have propelled dramatic decrease in unit cost PV panels and concentrated solar power (CSP) technology. Energy efficiency of the panels has also been improved. The combining effect have resulted in exponential decrease of cost per watt of solar power. In the year $2017$ the production of solar power has increased by $35\%$ world-wide, although having a global power share of $1.8\%$, the steady growth rate for past several years has projected solar power as the largest contributor of renewable energy in near-future.

However integrating solar energy to existing infrastructure brings major challenges. The traditional power infrastructure used load forecasting for efficient utility management, as there were random variability in the load and conventional power source had to produce a steady energy to compensate the same. After introduction of weather dependent source of energies like solar or wind in the grid, the weather introduced variability is significant on both demand and production side. Hence, in order forecast net demand of conventional energy at producer side, a load forecast must be accompanied by accurate prediction of the intermittent power\cite{hong2016probabilistic}. Predicting solar power availability at a particular time is also important for power trading market as cost of energy at real-time is often several times higher than day-ahead market; moreover solar energy producers are often penalized if supplied power is outside the tolerance interval of the committed power. Thus accurate solar forecasting has become a building block for power industry along with load forecasting.

Solar forecasting essentially have two parts, first is the forecasting of the weather variables like solar irradiation, wind speed, temperature etc. and the second is the prediction of the final power output or efficiency of a photo-voltaic panel. We will focus our efforts towards improving the former, as modeling device characteristics is separately handled given accurate prediction of weather parameters and age of the device. An extensive study of existing PV forecasting methods is present in  \cite{antonanzas2016review}.

There are mainly two measurements of solar irradiation, the direct normal irradiation (DNI) measures the irradiation coming from sun in a straight line, and direct horizontal irradiation (DHI) accounts to the irradiation due to scattering coming from other directions. Together they are measured by global horizontal irradiation (GHI), the amount of energy received at unit surface area. Forecasting GHI is divided into short-term and long-term forecasting according to the lead time. Short-term forecast predicts the irradiation from an hour ahead to an week ahead while long-term forecast generally tries to predict seasonal effects on the irradiation. While the short-term forecasting is important for utility management, long-term forecasting is more relevant for revenue generation and financial planning.

In this paper we show how sequence-to-sequence deep learning models compare with respect to the state of the art models in $\{1,..,24\}$ hour ahead solar irradiation forecasting.  Specifically, we implement encoder-decoder networks of Long Short Term Memory (LSTM), bidirectional LSTM, Recurrent Neural Network(RNN), GRU models and compare them with respect to existing Feed-forward neural network(FNN), Gradient Boosted Regression Trees(GBRT). We show that using state-of-art GBRT and FNN has certain short-coming which we can overcome using deep learning techniques. 

Additionally, we present deep learning models which uses meteorological data of $16$ neighborings locations, that resulted in significantly improving prediction accuracy for $[1-6]$ hour ahead predictions.

\section{Related Work}
Traditionally time series forecasting was under the scope of statistics and models like Auto Regression Integrated Moving Average (ARIMA), Seasonal ARIMA were extensively used linear models\cite{brockwell2002introduction}. But due to possible non-linearity in the data Artifical Neural Networks\cite{dolara2018comparison} were introduced and subsequently hybrid ARIMA models were used\cite{zhang2003time,ogliari2017physical}. Random Forest based methods were reported used in forecasting.\cite{nagy2016gefcom2014}, however the above were outperformed by Gradient Boosted Regression Tree (GBRT) while used for forecasting problem in  Global Energy Forecasting Competition 2014 \cite{hong2016probabilistic}. A summary of related PV power forecasting models is available in \cite{raza2016recent}. 

If additional weather data of neighboring locations are available, it can improve forecasting accuracy of a location. Features like wind speed, temperature  of the neighboring loactions were used in  Vector Auto Regression (VAR) and conditional VAR models, to improve the short-term wind forecast accuracy upto $6$ hours\cite{tastu2010multivariate,bessa2015probabilistic}. However for larger lead time spatial data do not improve accuracy as local weather data cease to become a determining factor for the same.

In deep learning encoder-decoder network using LSTM and RNN were first used for machine translation as a sequence-to-sequence learning problem, where length of the input and output sequence is not fixed. \cite{sutskever2014sequence,cho2014learning}. However it was then applied to forecasting problems where the lengths were known, yet they are not reported to be used in solar irradiation forecasting so far.\cite{isaksson2018solar}\cite{zaytar2016sequence}. In this paper we implement such model both with and without neighboring location.

\section{Data Source and Preparation}
\label{sec:Data Source}
National Renewable Energy Laboratory (NREL), US Department of Energy and Ministry of New and Renewable Energy (MNRE) India, have made available solar resource maps and related meteorological data on a $10$-kilometer grid across India. The data was captured through weather satellites in hourly basis from the year $2000$ to $2014$ and is available through National Solar Radiation Data Base (NSRDB)\cite{habte2017evaluation}. Experiments reported in the paper are based on Kharagpur, however relative performance of the models are not location specific, as experiments with other places given similar results. 

The forecasting models are trained on year $2000-2011$ data and tested on year $2012-2014$ data. Although the NSRDB data set carries several meteorological features, for single location based forecasting, only past GHI values of the target location were used to predict future GHI. However for multiple location based models wind speed, wind direction and GHI of each neighboring locations were used along with past GHI values of target location, in order to predict future GHI of target location. Below is a formal description of the problem setting:

\begin{figure}
\centerline{\includegraphics[width=0.4\textwidth]{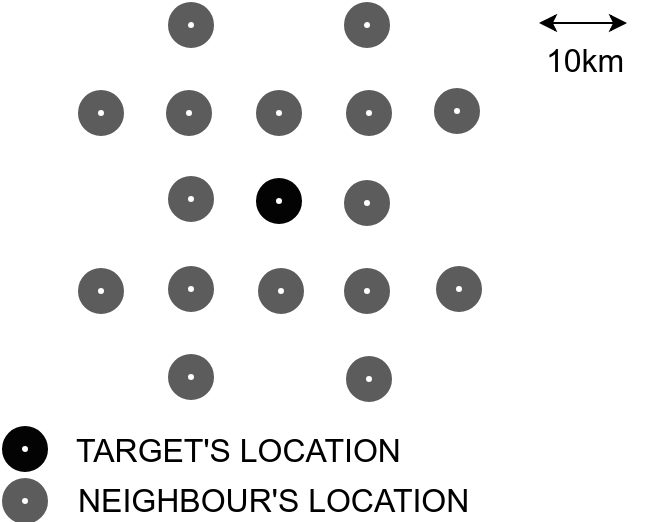}}
\caption{target location data with neighbours location.}
\label{fig1}
\end{figure}

\subsubsection{Single location based forecasting:}
Let GHI of the target location $j$ at time $t$ is denoted by $I_j(t)$. We want to predict the GHI of location $j$ after a time $T\in\{1,2,..,24\}$, called the lead time. In this setting  the future irradiation after time $T$, $I_j(t+T)$ is imagined as a function of past $p$ irradiations with respect to current time $t$, $\{I_j(t-p+1),..,I_j(t)\}$, where $p$ is the lag time, set to $120$ by validation.
 
 Hence training data is a pair $(X,y)$, where $X\in \mathbb{R}^{n \times p}, y \in \mathbb{R}^{n}$, and $t^{th}$ row  of $X$ is given by,

\begin{align*}
X[t,:]&=[I_j(t-p+1),..,I_j(t)] \\
y[t]&=I_j(t+T)
\end{align*}

Observe, for each value $T$ we have different model learned. Hence, length of the output sequence is always one, giving $I_j(t+T)$. Although this is not necessary for sequence-to-sequence models, it gives us flexibility to choose different kind model for different lead time.
 

 \subsubsection{Multiple location based forecasting} In this setting we take GHI values of $N$ neighboring locations along with the target location, since the future GHI of a target location is likely to have dependency on the recent past GHIs of the neighboring locations. The lag of GHI feature for target location is $p$ and that of the neighboring locations is $p'$. We do not use same lag for the neighboring locations because that would result in a heavier model which would require more data to train. However, we additionally use current hour wind speed and wind direction of all $N+1$ locations as input. Hence each sequence of input data has a dimension $d=p+Np'+2(N+1)$, while the output is a scaler $I_j(t+T)$, GHI of the target location $j$ after time $T$.
 
Let, $W_k(t)$ and $S_k(t)$ denote the wind direction and wind speed of location $k$ at the time $t$, then training data can be represented as a pair $(X,y)$, where $X\in \mathbb{R}^{n \times d}, y \in \mathbb{R}^{n}$, and $t^{th}$ row  of $X$ is given by,

\begin{align*}
X[t,1:p]&=[I_j(t-p+1),..,I_j(t)] \\
X[t,p+(k-1)p':p+kp']&=[I_k(t-p'+1),..,I_k(t)]\\
X[t,p+Np'+k]&=W_k(t) \quad \forall k\in [N]\cup \{j\}\\
X[t,p+N(p'+1)+k+1]&=S_k(t) \quad\forall k\in [N]\cup \{j\}\\
y[t]&=I_j(t+T)
\end{align*}

The intuition behind using GHI, wind speed and wind direction of the neighboring locations is that a sudden drop in GHI at a neighboring location due to cloud cover, would affect the target location if the wind direction is towards it. Figure \ref{fig1} shows the choice of $N=16$ neighbors in our experiments. 
 

%

\section{ Model Design} 
\begin{figure}
	\centerline{\includegraphics[width=0.4\textwidth]{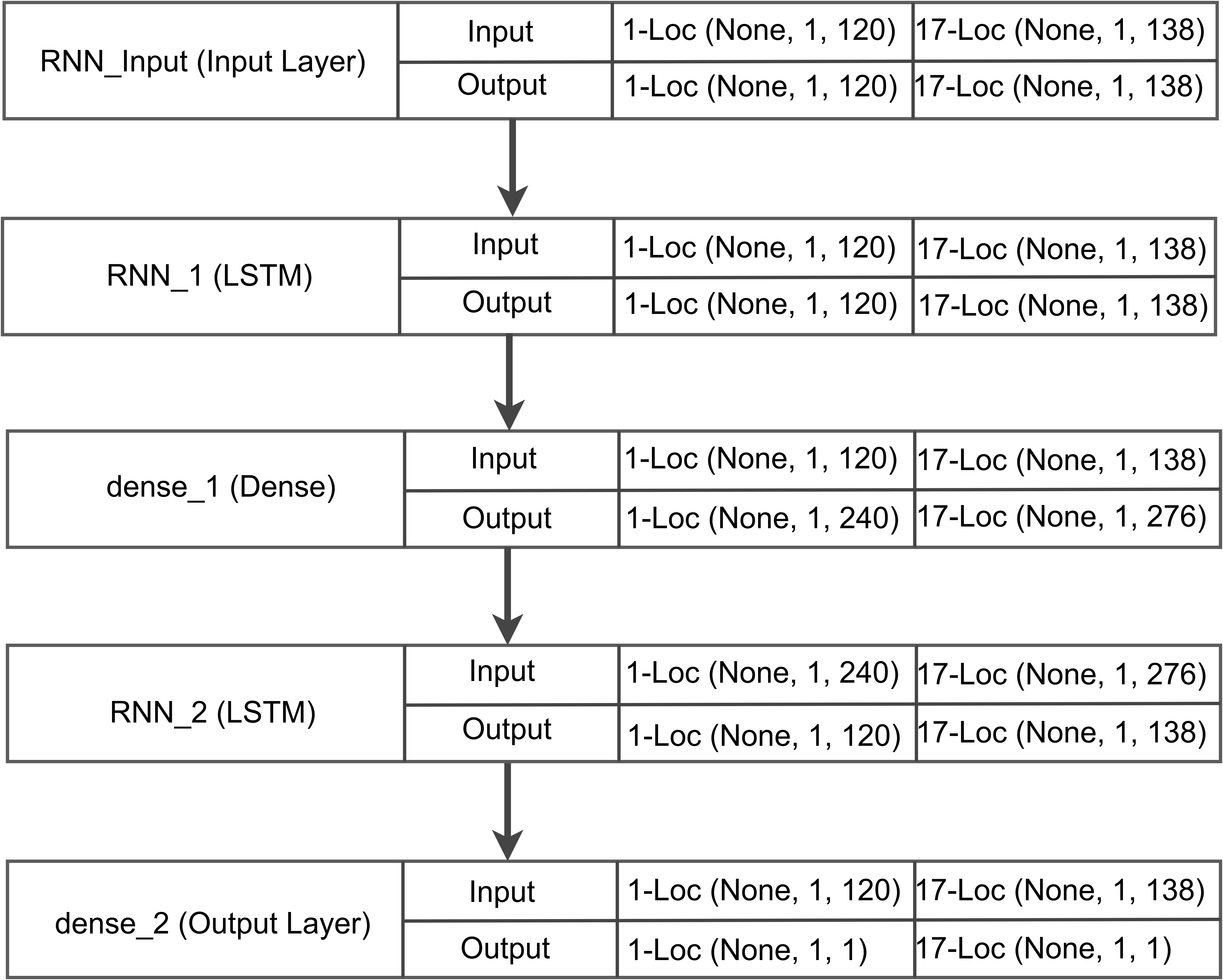}}
	\caption{RNN encoder-decoder network for single and multi location}
	\label{fig:rnn}
\end{figure}

\subsection{RNN and LSTM model architecture:}
Recurrent Neural Networks(RNN) are generalized neural network with a memory cell which is updated at each time step. This enables simple RNNs with a short term memory. To cater long term dependency in the input and output of the network Long Short Term Memory (LSTM) were introduced\cite{hochreiter1997long}. However for a machine translation problem, the length of input and output sequence does not remain fixed, so an encoder network translates the variable length input to a fixed length representation, which is then translated back to a variable length output sequence\cite{cho2014learning}. However encoder-decoder architectures are also used to capture long term dependency in the data, which is common in time series forecasting, although here the length of the sequences remains fixed. 

We first construct encoder-decoder networks with simple RNN as encoder and decoder with a dense layer in between them to capture the fixed representation. The final single dimensional output goes through a dense layer since the GHIs have to be appropriately scaled. For both the single and multiple location we used the same architecture, but the number of RNN units changes according to the input dimension. Figure \ref{fig:rnn}, describes both model in a single diagram. For all layer except final layer we have used activation function Scaled Exponential Linear Units (SELU), which has self normalizing property and robustness to outliers\cite{klambauer2017self}. In the final dense layer we have used traditional Rectified Linear Unit (ReLU) as activation function\cite{nair2010rectified}.

  For single location based model, the input sequence is $p=120$ dimensional, for multiple location it is $d=p+Np'+2(N+1)=138$ dimensional, given lag for target location $p=72$, lag for neighboring location $p'=2$ and total number of neighbors $N=16$. Observe that the representation layer $dense\_1$ has dimension twice than that of the input layer, such a architecture is found in visual network of flies, the higher dimesional representation helps in generalization.

\begin{figure}
\centerline{\includegraphics[width=0.4\textwidth]{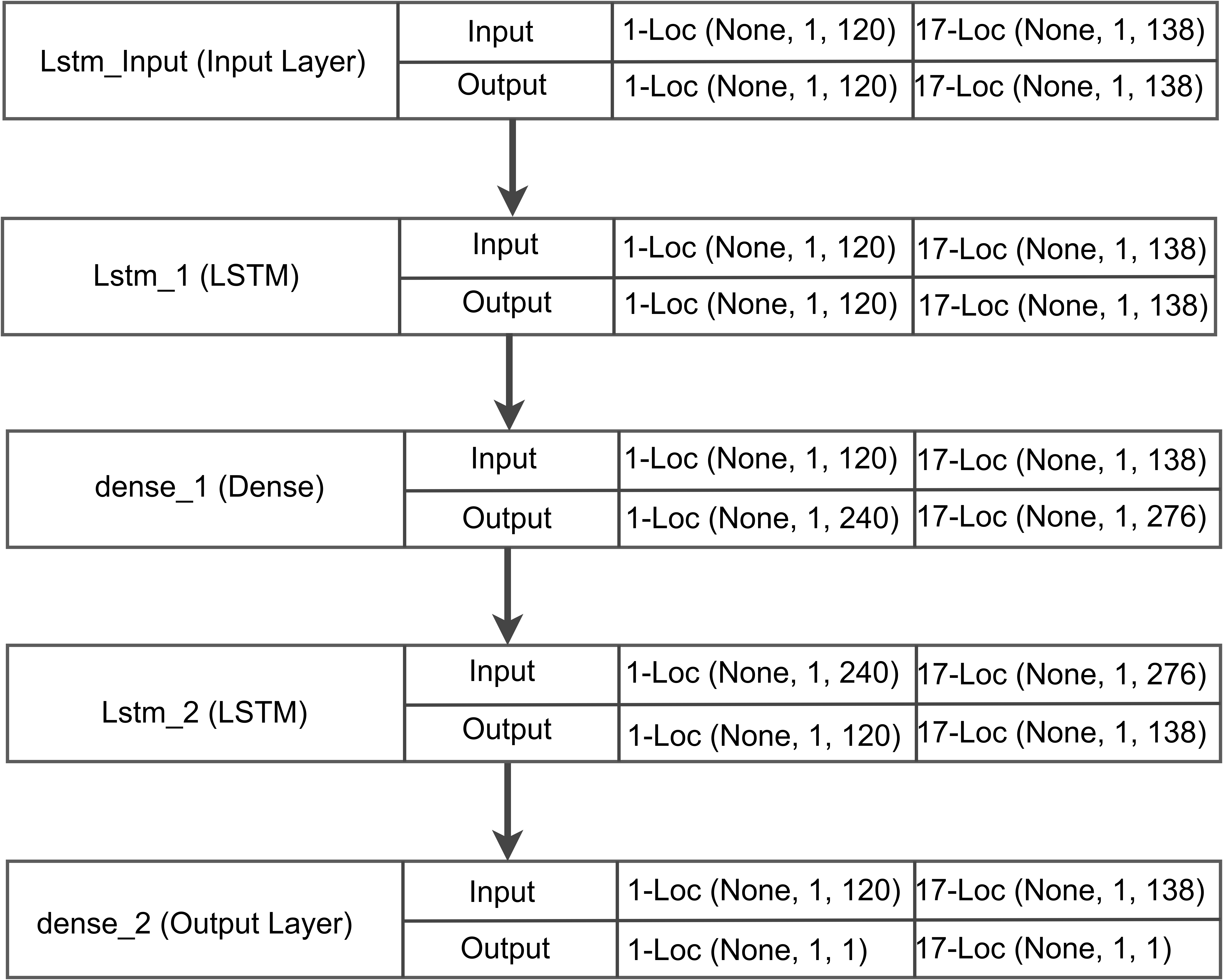}}
\caption{LSTM encoder-decoder network for single and multi location}
\label{fig:lstm}
\end{figure}

The LSTM encoder-decoder architecture is similar, only LSTMs are used instead of RNNs as encoder and decoder. Figure \ref{fig:lstm} described the combined architecture of single and multi location based model as before. Models are trained with MAE loss function and Adam\cite{kingma2014adam} optimizer upto maximum 100 epochs.

\subsection{Feed Forward Neural Nets:} 
\begin{figure}
\centerline{\includegraphics[width=0.4\textwidth]{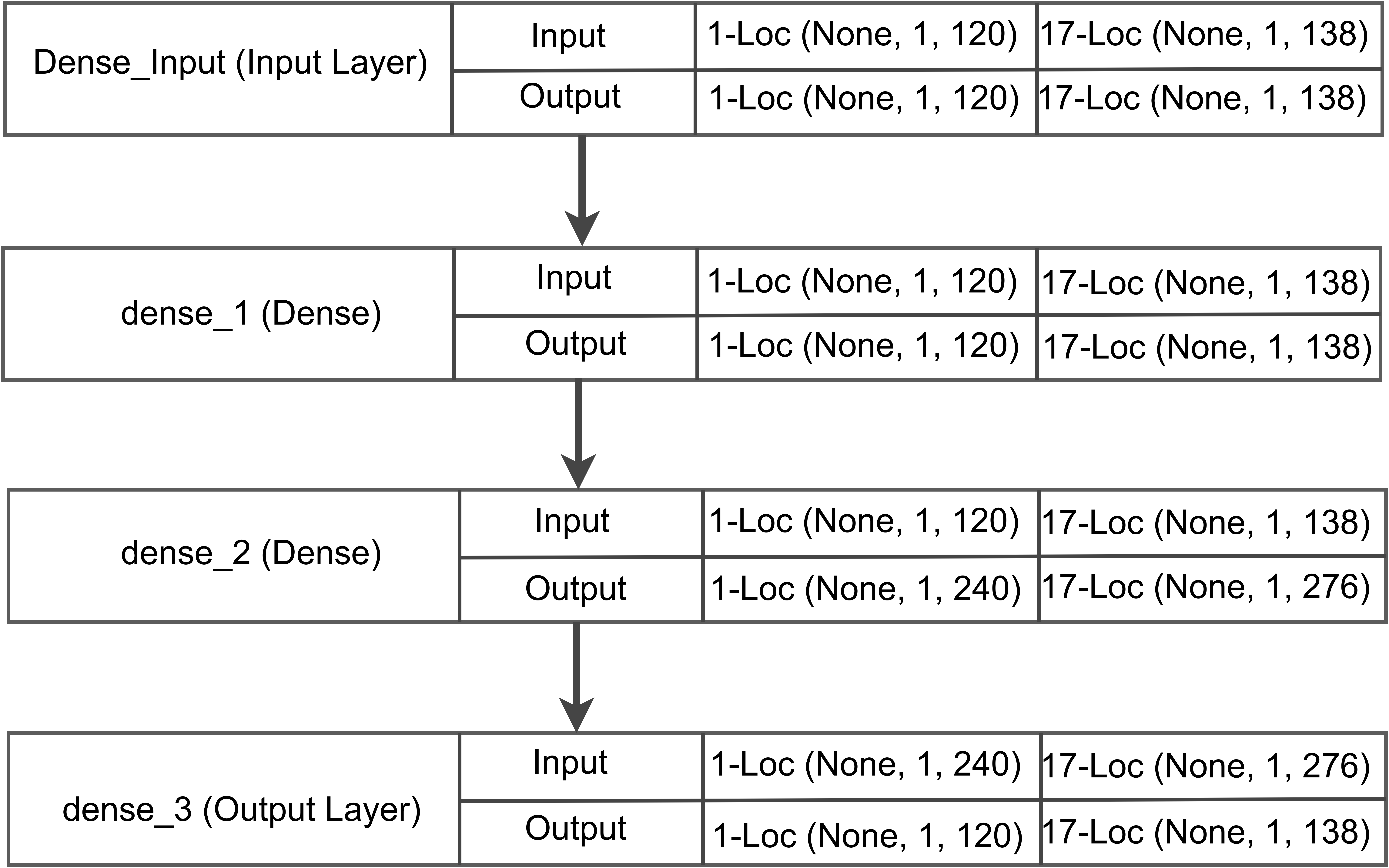}}
\caption{FFNN for single and multi location based forecasting.}
\label{fig:ffnn}
\end{figure}

FFNNs are common for classification and regression task. For the task at hand we have stacked $3$-dense layers, where the two hidden layers have dimension twice than the input layer. The final dense layer uses activation function ReLU, while hidden layes use SELU\cite{klambauer2017self}. Experiments with deeper layers does not improve the forecasting accuracy. Figure \ref{fig:ffnn} shows our the architecture with both single and multi location together. Models are trained with MAE loss function and Adam\cite{kingma2014adam} optimizer upto maximum 100 epochs.

\subsection{Gradient Boosted Regression Trees:} Adaptive Boosting(AdaBoost) algorithm was introduced as an ensemble of method, where successive models were introduced to decrease the loss by re-weighting the data points\cite{freund1996experiments}. Later it was recognized as a gradient descent on a special loss function and successively generalized for other loss functions as gradient boosting algorithm\cite{friedman2000additive,friedman2001greedy}. It works as an ensemble of models where successive weak learners are introduced to minimize existing errors in the training set. Gradient boosting on regression trees immediately propossed\cite{mason2000boosting}.

In the experiments we have used Huber loss function for training, as using MAE as training loss had very high RMSE. Huber is strikes a balance between these two losses, as it treats higher residual linearly and smaller residuals quadratically. The maximum tree depth was set to $6$ for single location based models and $8$ multiple location based models, as to minimize validation error.

\section{Results and Discussion}

Mean Absolute Error (MAE) and Root Mean Squared Error (RMSE) are standard loss functions for quantifying model test performance for regression. For a target location, let $y[t]$ be actual GHI and $\hat{y}[t]$ be predicted GHI, then the metrics are defined as follows:

\begin{align*}
MAE &= \frac{1}{n}\sum\limits_{t=1}^{n} \lvert y[t]-\hat{y}[t] \rvert\\
RMSE &= \sqrt{\frac{1}{n}\sum\limits_{t=1}^{n} (y[t]-\hat{y}[t])^2}
\end{align*}

however in presence of corrupted data or outliers, RMSE is dominated by large deviations as errors get squared. As solar irradiation data often gets corrupted because of sensor errors or sudden change in weather, it is usually quantified using MAE, to reduce it's susceptibility to outliers.

	\begin{center}
		
		\begin{table}
			
			\begin{tabular}{ |p{1.3cm}|p{0.75cm}|p{0.75cm}|p{0.75cm}| p{0.75cm}| p{0.75cm}| p{0.75cm}|}
				\hline
				\multicolumn{1}{|c|}{}& \multicolumn{2}{|c|}{T= 1 hr} & \multicolumn{2}{c|}{T=24 hr} & \multicolumn{2}{c|}{24 hr avg} \\\cline{2-7}
				Model &MAE& RMSE&  MAE & RMSE & MAE & RMSE\\
				\hline
				FFNN-1   &21.3 & \textbf{57.9} &32.4 & 86.0 & 31.1& 81.1\\
				RNN-1 & 20.9&  59.3&32.7 &87.4&30.9&83.0\\
				LSTM-1 &\textbf{20.6} &58.5 &\textbf{31.9} &85.3&\textbf{30.3}&81.1\\
				GBRT-1     &21.5 &58.0 & 32.7	 & \textbf{82.9} & 31.3 & \textbf{79.2}\\
				\hline
				FFNN-17 &17.8 &\textbf{49.6} & \textbf{31.5} & 84.7 & 30.0 &80.2	\\
				RNN-17  &\textbf{17.7} &\textbf{49.6} & 32.4& 83.8 & \textbf{29.9}& 80.0\\
				LSTM-17 &18.1 &50.3 & 31.7& 83.8& \textbf{29.9} & 80.0\\
				GBRT-17 &19 &50.7 & 31.9& \textbf{81.3}& 30.9& \textbf{78.2}\\
				\hline
			\end{tabular}
			\caption{}
			\label{table:error}
			
		\end{table}

	\end{center}
	
We compare FFNN and GBRT methods with LSTM and RNN encoder-decoder for forecasting future irradiation upto $24$-hour for both single-location (FFNN-1, RNN-1, LSTM-1, GBRT-1) and multi-locaton (FFNN-17, RNN-17, LSTM-17, GBRT-17) based models. Here $17$ stands for the total number of locations consisting one target location and $16$ neighboring locations. Table \ref{table:error} reports the MAE and MSE of the models in $W/m^2$, for lead time $T=1$, $T=24$ and a average taken over all $T\in \{1,2,..24\}$. The minimum error in each setting is highlighted in bold. 

\begin{figure}
\centerline{\includegraphics[width=9.5cm, height=7cm]{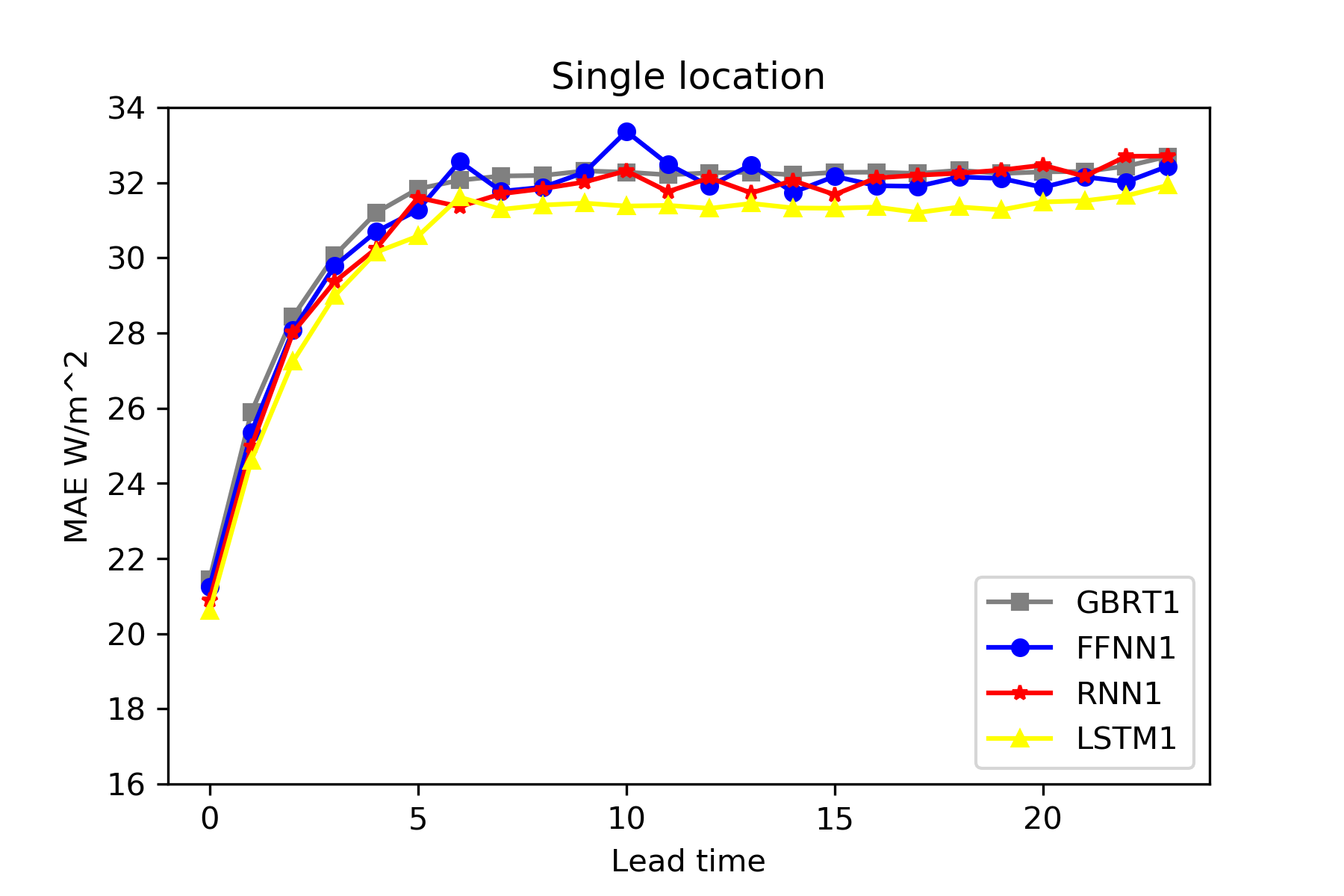}}
\caption{Figure shows MAE of forecasted GHI for single location based models for lead time T=1 to 24 hours. It can observed that LSTM performs the best for all 24 hours, with average error 30.3 $W/m^2$}
\label{fig:mae_1}
\end{figure}

\begin{figure}
\centerline{\includegraphics[width=9.5cm, height=7cm]{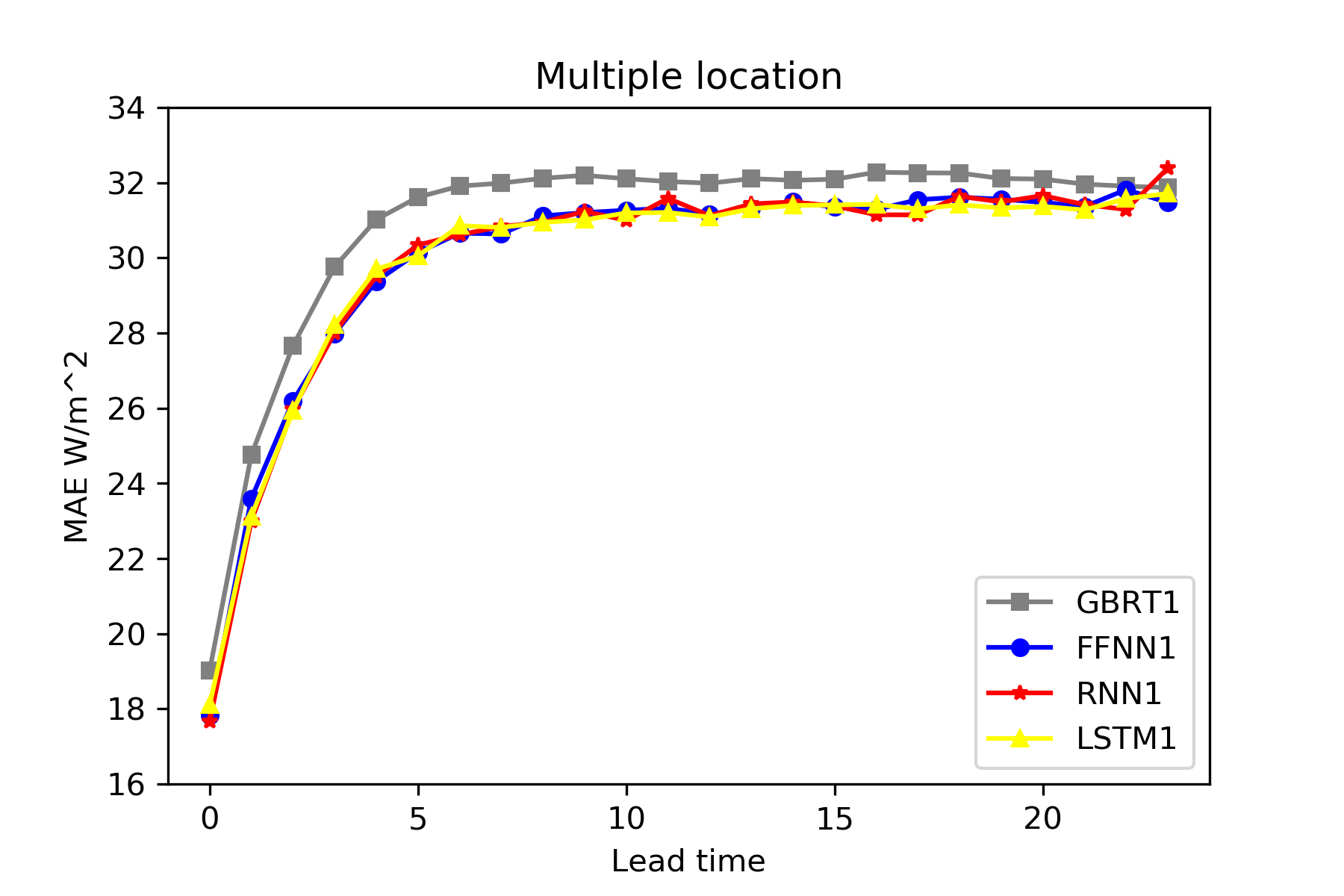}}
\caption{Figure shows MAE of forecasted GHI for multiple location based models for lead time T=1 to 24 hours. It can be observed that LSTM, RNN performs considerably better with average error 29.9 $W/m^2$, however GBRT has higher average error 30.9 $W/m^2$}
\label{fig:mae_17}
\end{figure}

The average performance indicates that for single location experiments LSTM minimizes the MAE ($30.3\, W/m^2$) better than other models. The same can be visualized in figure \ref{fig:mae_1}, where LSTM is better than other models for all 24 hour. A similar trend is there for multiple locations, where LSTM and RNN both gives minimum MAE ($29.9\, W/m^2$). Figure \ref{fig:mae_17} shows their detailed performance. 

The advantage of using multi-location based models over single locations do not look significant when average performance over 24 hours are observed, however multi-location based model reduces GHI errors by minimum $11\%$ to maximum $16\%$ for 1 hour ahead prediction, across the models under consideration.

\begin{figure}
	\centerline{\includegraphics[width=9.5cm, height=7cm]{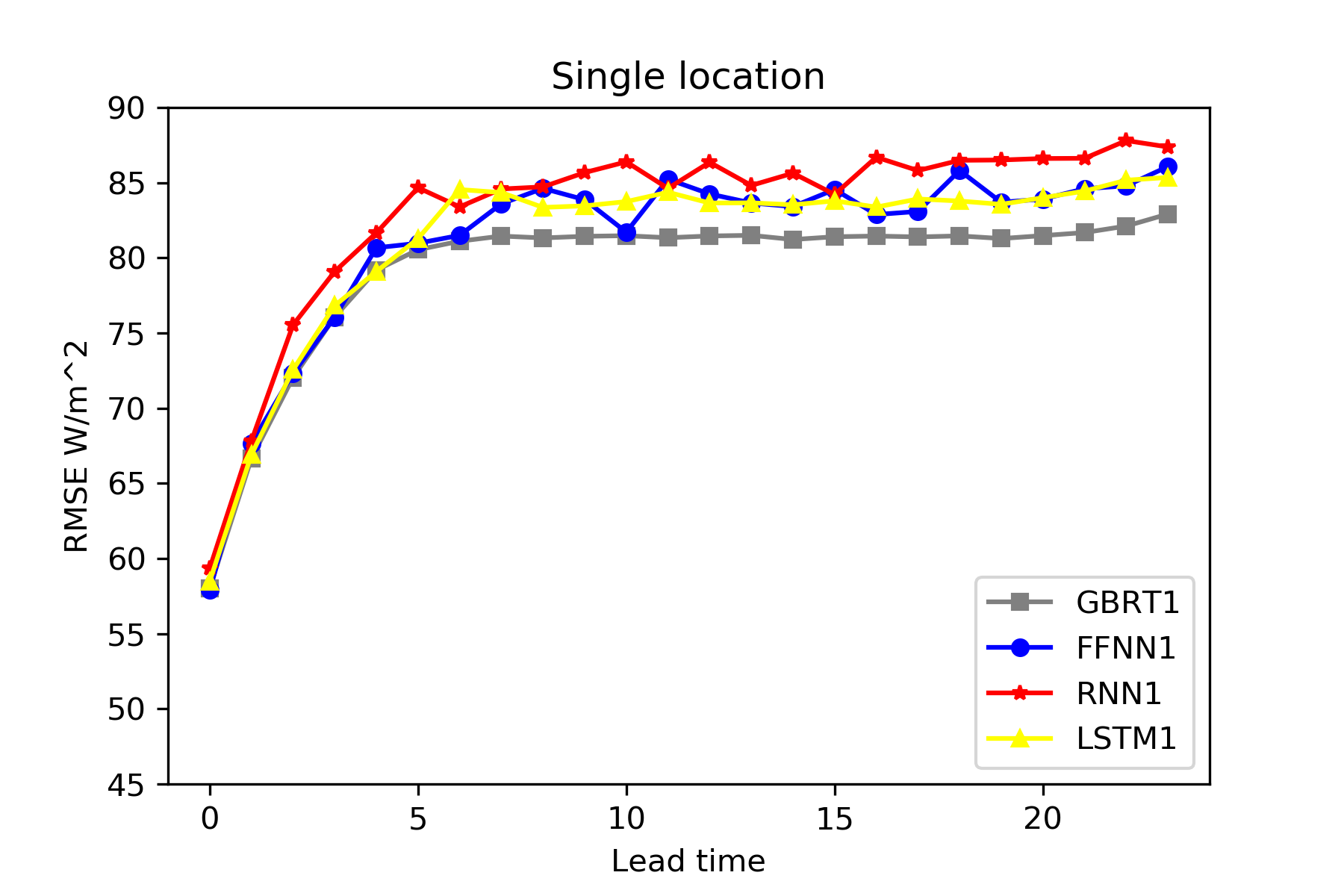}}
	\caption{Figure shows RMSE of forecasted GHI for single location based models for lead time T=1 to 24 hours. It can observed that GBRT performs the better over 24 hours, with average error 79.2 $W/m^2$}
	\label{fig:rmse_1}
\end{figure}

\begin{figure}
	\centerline{\includegraphics[width=9.5cm, height=7cm]{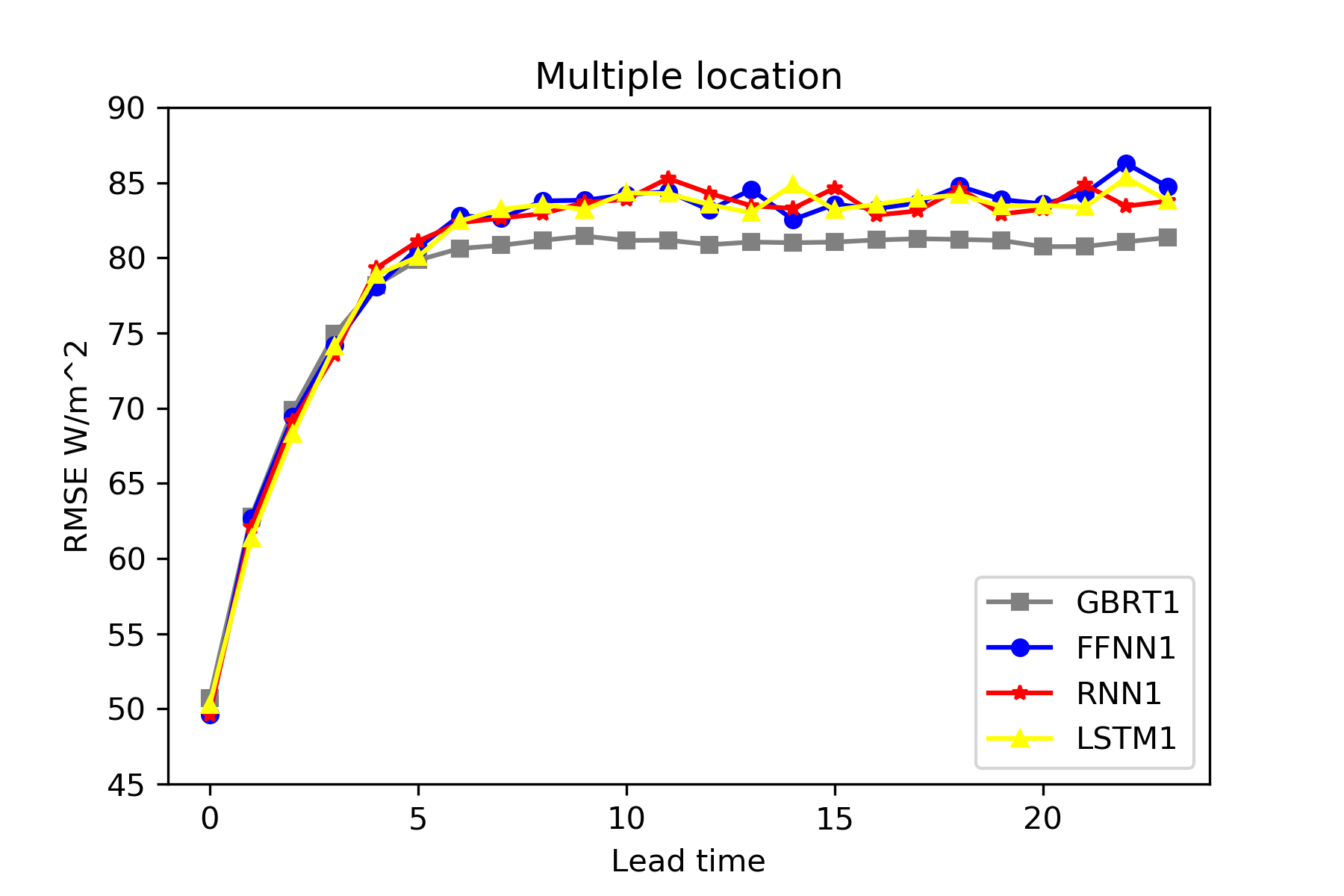}}
	\caption{Figure shows RMSE of forecasted GHI for multiple location based models for lead time T=1 to 24 hours. It can observed that GBRT performs the better over 24 hours, with average error 78.2 $W/m^2$}
	\label{fig:rmse_17}
\end{figure}

With respect to RMSE, GBRT gives minimum average error for both single ($79.2\, W/m^2$) and multi location ($78.2\, W/m^2$) based models. Figure \ref{fig:rmse_1} and \ref{fig:rmse_17} depicts their performance for all 24 hours, and it's superior performance over other models justifies it's place current literature. It can be seen than for initial hours, FFNN has slight edge over GBRT with respect to RMSE.

However it is worth re-iterating, that if GBRT is trained using MAE loss, it results smaller MAE but very high RMSE, which forces us to choose Huber as training loss. 

\section{Conclusion}

We introduced deep encoder-decoder models for solar irradiation forecasting and compare it with state-of-art models. It was found that LSTM models are suitable to minimize MAE loss, also they don't have very high RMSE error ($2.5\%$ higher). The experimental result also shows that GBRT is not suitable for reducing both MAE and RMSE loss simultaneously. We additionally give the multi-location based deep learning models, and show performance improvement over single location based models.

\section*{Acknowledgment}
The authors would like to thank Dr. Purushottam Kar, Indian Institute of Technology Kanpur for his valuable suggestions in setting up the models. This work is supported by UK-India Clean Energy Research Institute project jointly funded by Department of Science and Technology (DST), Govt. of India and Engineering and Physical Sciences Research Council (EPSRC), UK 
\bibliography{ghi_ref} 
\bibliographystyle{unsrt}

\end{document}